\documentclass[journal]{IEEEtran}
\ifCLASSINFOpdf
\else
\fi

\usepackage{graphicx}
\usepackage{times}
\usepackage{soul}
\usepackage{url}
\usepackage{subfigure}
\usepackage{caption}
\usepackage[hidelinks]{hyperref}
\usepackage[utf8]{inputenc}
\usepackage{amsmath,amssymb}
\usepackage{booktabs}
\usepackage{algorithm}
\usepackage{algorithmic}
\usepackage{mathtools}
\usepackage{array}
\usepackage{multirow}
\usepackage{stmaryrd}
\usepackage[misc]{ifsym}
\usepackage{booktabs}
\usepackage{tabularx}
\usepackage{xcolor}
\usepackage{pdflscape}

\begin{document}
\title{Source-Free Cross-Domain Continual Learning}
% \author{Anonymous Authors}

\author{M.T.~Furqon, Mahardhika~Pratama, Igor~ŠKRJANC, Lin~Liu, Habibullah~Habibullah
       and~Kutluyil~Dogancay% <-this % stops a space
\thanks{M. T. Furqon, Mahardhika Pratama, Lin Liu, Habibullah Habibullah and Kutluyil Dogancay are with the STEM, University of South Australia, Mawson Lakes Boulevard, Adelaide,
Australia, 5095.}% <-this % stops a space
\thanks{Igor ŠKRJANC is with the Faculty of Electrical and Computer Engineering, University of Ljubljana, Slovenia.}% <-this % stops a space
%\thanks{Manuscript received April 19, 2005; revised August 26, 2015.}%
}


\markboth{Journal of \LaTeX\ Class Files,~Vol.~14, No.~8, August~2015}%
{Shell \MakeLowercase{\textit{et al.}}: Bare Demo of IEEEtran.cls for IEEE Journals}



\maketitle

\begin{abstract}
Although existing cross-domain continual learning approaches successfully address many streaming tasks having domain shifts, they call for a fully labeled source domain hindering their feasibility in the privacy constrained environments. This paper goes one step ahead with the problem of source-free cross-domain continual learning  where the use of source-domain samples are completely prohibited. We propose the idea of rehearsal-free frequency-aware dynamic prompt collaborations (REFEREE) to cope with the absence of labeled source-domain samples in realm of cross-domain continual learning. REFEREE is built upon a synergy between a source-pre-trained model and a large-scale vision-language model, thus overcoming the problem of sub-optimal generalizations when relying only on a source pre-trained model. The domain shift problem between the source domain and the target domain is handled by a frequency-aware prompting technique encouraging low-frequency components while suppressing high-frequency components. This strategy generates frequency-aware augmented samples, robust against noisy pseudo labels. The noisy pseudo-label problem is further addressed with the uncertainty-aware weighting strategy where the mean and covariance matrix are weighted by prediction uncertainties, thus mitigating the adverse effects of the noisy pseudo label. Besides, the issue of catastrophic forgetting (CF) is overcome by kernel linear discriminant analysis (KLDA) where the backbone network is frozen while the classification is performed using the linear discriminant analysis approach guided by the random kernel method. Our rigorous numerical studies confirm the advantage of our approach where it beats prior arts having access to source domain samples with significant margins. Source codes of REFEREE are made publicly available in \url{https://github.com/furqon3009/REFEREE}.

% \url{https://anonymous.4open.science/r/REFEREE-FEE0}.

\end{abstract}


% \fi
\IEEEpeerreviewmaketitle

\section{Introduction}
\IEEEPARstart{T}{he} goal of continual learning (CL) is to deal with lifelong learning environments where a sequence of non-stationary tasks is observed. One of the major challenges is the catastrophic forgetting (CF) problem \cite{chen2018lifelong,Parisi2018ContinualLL,Masana2020ClassIncrementalLS,Zhou2023ClassIncrementalLA} resulting in significant performance drops of previous tasks because previously valid parameters are over-written with new ones when learning a new task with the absence of any old samples. On the other hand, the naive retraining solution is undesired because of computational and memory constraints. This issue has triggered an active study where the goal is to balance the stability-plasticity aspect \cite{Kim2023OnTS} such that a learning agent is capable of mastering all seen tasks ultimately. 

There exist three common approaches for CL \cite{Parisi2018ContinualLL}: memory-based approach, regularization-based approach and architecture-based approach. The memory-based approach \cite{lopez2017gradient,Chaudhry2018EfficientLL,Chaudhry2019UsingHT,Buzzega2020DarkEF,Dam2022ScalableAO,Masum2023AssessorGuidedLF,Rebuffi2016iCaRLIC,VinciusdeCarvalho2022ClassIncrementalLV,Shin2017ContinualLW} stores old exemplars to be interleaved with new samples when learning a new task. Nevertheless, this approach imposes high memory complexities and class imbalances if the memory size is kept bounded. The regularization-based approach \cite{aljundi2018memory,Kirkpatrick2016OvercomingCF,Li2016LearningWF,Paik2019OvercomingCF,Cha2020CPRCR,Schwarz2018ProgressC,Zenke2017ContinualLT} inserts a regularization term in the loss function penalizing variations of important parameters of old tasks. Although this approach is easy to implement, the performance is relatively lower than the memory-based approach. In addition, this approach does not scale well to a large number of tasks because of the difficulty in finding overlapping regions of these tasks. The architecture-based approach \cite{Li2019LearnTG,Pratama2021UnsupervisedCL,Ashfahani2021UnsupervisedCL,Rusu2016ProgressiveNN,Rakaraddi2022ReinforcedCL,Yoon2017LifelongLW} relies on a network expansion strategy while isolating old network parameters. This approach is over-dependent on task ID during the inference process. Although the CL area has rapidly progressed, the vast majority of existing works are limited to a single CL process and not transferable to different but related CL processes.

Such issue has led to the advent of cross-domain continual learning \cite{VinciusdeCarvalho2024TowardsCC,Lao2021ATC,Weng2024CrossDomainCL} featuring the domain shift problem between source and target domains. That is, it aims to develop a domain-invariant model to perform knowledge transfer from the label-rich source domain to the unlabeled target domain. It goes one step ahead of traditional domain adaptation paradigms \cite{Ganin2015DomainAdversarialTO} where both source and target domains feature streaming tasks, e.g., sequence of tasks. \cite{Lao2021ATC} handles this problem with the notion of generative replay, while \cite{VinciusdeCarvalho2024TowardsCC} proposes the idea of cross-attention. \cite{Weng2024CrossDomainCL} offers the concept of meta learning to overcome the cross-domain continual learning problem. \textit{The cross-domain continual learning remains an open research area because they depend on a strong assumption that the labeled samples of the source domain are accessible when performing knowledge transfer}. Such assumption might be impractical because of the privacy and storage concerns.   

This paper puts forward for the first time the source-free cross-domain continual learning (SFCDCL) problem disallowing the use of labeled source-domain samples for continual domain adaptations to comply with the privacy requirements. Only a pre-trained source model is given to perform continual domain adaptations. Note that our problem differs from \cite{Lin2022PrototypeGuidedCA} where a source domain also features streaming tasks. That is, the pre-trained source model does not gain access to all classes before the process commences, pronouncing the CF problem to some extents. The SFCDCL features three major challenges. The first challenge, namely over-dependence on the pre-trained source model, occurs in the existing source-free domain adaptation and causes suboptimal generalization. In other words, this limitation leads to the use of the same model as that of the source domain. The second challenge lies in the noisy pseudo-labels because the SFCDCL problem solely relies on unlabeled target-domain samples. Given the presence of the domain shift problem between the labeled source domain and the unlabeled target domain, the risk of noisy pseudo labels generated by the pre-trained source model is high. This problem is further aggravated by the fact that a deep neural network suffers from the over-confidence problem. Since the source domain and the target domain present sequences of tasks, it opens the third challenge, double catastrophic forgetting (DCF), where the CF problem might ensue in both source domain and target domain. The DCF challenge calls for not only the training process of the target domain to be protected against the CF problem but also that of the pre-training process of the source domain. 

To this end, this paper proposes a rehearsal-free frequency-aware dynamic prompt collaboration (REFEREE). REFEREE presents a dual-branch network structure implementing a collaborative training strategy between the source-pre-trained model and the vision-language model consequently addressing the challenge 1, namely the over-dependence on the pre-trained source model. Such approach enables extractions of domain-specific information while consolidating domain-invariant knowledge \cite{Zhan2024TowardsDC}. Specifically, the domain-invariant knowledge is extracted by the frozen vision-language model while the domain-specific knowledge is captured by the prompted source-pretrained model, ViT \cite{Dosovitskiy2020AnII}. The challenge 2 is tackled with the notion of frequency-aware prompting technique. Since it is well known that high-frequency images are susceptible to noises, the high-frequency images are suppressed during the training process while promoting the low-frequency images. Firstly, the images are decomposed in the frequency domain where the high-frequency components are converted either to normally distributed values or zeros. In other words, the high-frequency components are modified and result in augmented samples. Secondly, these samples are further learned in the adversarial manner with the mutual attention strategy \cite{Zhang2024ExploringCF} as with the task augmentation strategy of \cite{Wang2021CrossDomainFC}. In addition, the noisy pseudo label problem is dealt with the uncertainty-aware weighting scheme. That is, uncertainty weights are derived based on the Shannon entropy and in turn applied to weight the mean and covariance matrix, thereby minimizing the detrimental effects of the noisy pseudo label problem. The DCF problem is overcome by the kernel linear discriminant analysis (KLDA) technique inspired by \cite{Momeni2024ContinualLU}. That is, the backbone network is completely fixed during the training process to prevent the DCF problem. The classification is performed using the linear discriminant analysis method in both source and target domains. Nevertheless, the direct use of LDA results in suboptimal performances due to the linear decision boundary. This problem can be tackled using the kernel method but the construction of the kernel matrix is intractable in the continual learning context. It is approximated using the random Fourier features here. In other words, the class mean and the shared covariance matrix are calculated using the kernelized features leading to the LDA method enhancing the class separability. Our approach is completely gradient-free where the training process is done only to generate the mean and covariance matrix for inferences leaving the backbone networks frozen.     

This paper conveys at least three contributions:
\begin{itemize}
    \item This paper presents an uncharted problem, source-free cross-domain continual learning (SFCDCL) problem extending the CDCL problem with the absence of the labeled source-domain samples for domain adaptations. 
    \item This paper proposes the rehearsal-free frequency-aware dynamic prompt collaboration (REFEREE) to cope with the SFCDCL problem tackling the three research challenges: over-dependence on the pre-trained source model, noisy pseudo-labels, double catastrophic forgetting.   
    \item Extensive experiments are undertaken where REFEREE is compared with prior arts in several benchmark problems. REFEREE is capable of outperforming benchmarked algorithms with significant margins with the absence of any labeled source-domain samples.  
\end{itemize}

The remainder of this paper is structured as follows: Section 2 outlines related works, Section 3 discusses preliminaries, Section 4 details our method, REFEREE, Section 5 describes our experiments and Section 6 elucidates concluding remarks and future works. 

\section{Related Works}
\subsection{Continual Learning (CL)}
The CL topic is a rapidly growing research area to develop a solution for non-stationary and never-ending learning environments. Numerous works are recently devoted to study this topic where the goal is to minimize the catastrophic interference between new and past knowledge. The regularization-based approach \cite{aljundi2018memory,Kirkpatrick2016OvercomingCF,Li2016LearningWF,Paik2019OvercomingCF,Cha2020CPRCR,Schwarz2018ProgressC,Zenke2017ContinualLT} inserts a regularization term to protect important parameters of old tasks from deviations. The architecture-based approach \cite{Li2019LearnTG,Pratama2021UnsupervisedCL,Ashfahani2021UnsupervisedCL,Rusu2016ProgressiveNN,Rakaraddi2022ReinforcedCL,Yoon2017LifelongLW} relies on the parameter isolation strategy combined with a network expansion approach for plasticity. The memory-based approach \cite{lopez2017gradient,Chaudhry2018EfficientLL,Chaudhry2019UsingHT,Buzzega2020DarkEF,Dam2022ScalableAO,Masum2023AssessorGuidedLF,Rebuffi2016iCaRLIC,VinciusdeCarvalho2022ClassIncrementalLV,Shin2017ContinualLW} is underpinned by the experience-replay strategy using old exemplars stored in the memory. Recently, the CL area is advanced by the use of prompts with a pre-trained model, ViT \cite{Wang2021LearningTP,Wang2022DualPromptCP}. This concept promotes significant improvements while being computationally light compared to the previous three approaches. Notwithstanding that the CL area has enjoyed significant progresses, existing works mainly focus on a single CL process and do not allow for seamless knowledge transfer to different-but related CL process. In other words, each CL process is independently dealt with, thus incurring considerable complexities. 

\subsection{Source Free Domain Adaptation (SFDA)}
SFDA extends unsupervised domain adaptation (UDA) where a privacy constraint is introduced. That is, no labeled source domain samples are used for domain adaptations. Only unlabeled target domain samples and a pretrained source model are shared for domain adaptations. \cite{Liang2020DoWR} proposes a self-training mechanism using the cluster structure while \cite{Li2020ModelAU} applies a generative model to address the absence of source domain samples. \cite{Chen2022ContrastiveTA} puts forward the concept of self-supervised learning, \cite{Karim2023CSFDAAC} designs a curriculum learning strategy, \cite{Litrico2023GuidingPW} presents a loss-re-weighting strategy. Nonetheless, existing SFDA works merely consider a single task setting ignoring any continual learning requirements. 

\subsection{Cross-Domain Continual Learning (CDCL)}
CDCL combines the CL problem and the UDA problem overcoming the CL limitation where only a single CL process is considered at any given time. It is akin to the UDA setting where there exists the presence of the labeled source domain and the unlabeled target domain and the goal is to generalize well to the unlabeled target domain using label information of the source domain. The unique aspect lies in the streaming nature of each domain, i.e., each domain features a sequence of task. The key research question of the CDCL problem lies in how to cope with the domain shift problem and the CF problem simultaneously. \cite{Lao2021ATC} proposes the idea of generative replay to cope with the double catastrophic forgetting (DCF) problem, \cite{VinciusdeCarvalho2024TowardsCC} designs a rehearsal-based approach of compact convolution transformer, while \cite{Weng2024CrossDomainCL} applies the meta-learning principle to handle the domain shift and the CF problem.  However, the CDCL problem does not yet take into account the issue of privacy where the access of source-domain samples are prohibited. The SFCDCL problem goes one step ahead where the training process is supported only by a pretrained source model and unlabeled target-domain samples. It distinguishes itself from \cite{Yang2024NavigatingCT} where the pre-trained source model does not enjoy any access of all source domain classes. That is, the source domain also characterizes streaming tasks causing the double catastrophic forgetting (DCF) problem.   

\section{Preliminaries}
\subsection{Problem Formulation}
SFCDCL problem is defined as a learning problem of sequentially arriving tasks in the source domain $\mathcal{D}_{1}^{s},\mathcal{D}_{2}^{s},...,\mathcal{D}_{T}^{s}$ and in the target domain $\mathcal{D}_{1}^{tr},\mathcal{D}_{2}^{tr},...,\mathcal{D}_{T}^{tr}$ where the goal is to build a predictive model generalizing well in the target domain. $\mathcal{D}_{t}^{s}=\{(x_i,y_i)\}_{i=1}^{N_{t}^{s}}$ stands for the $t-th$ task of the source domain having full labels while $\mathcal{D}_{t}^{tr}=\{(x_{i})\}_{i=1}^{N_{t}^{tr}}$ denotes the $t-th$ task of the target domain suffering from the absence of any labels. $x_{i}\in\mathcal{X},y_{i}\in\mathcal{Y}$ are the $i-th$ image and its corresponding label where the source domain and the target domain are different but related $\mathcal{X}_{s}\times\mathcal{Y}_{s}\in\mathcal{S}$, $\mathcal{X}_{tr}\times\mathcal{Y}_{tr}\in\mathcal{T}$. That is, they are closed set or share the same target variables $\mathcal{Y}_{s}=\mathcal{Y}_{t}$ but are drawn from different domains $\mathcal{S}\neq\mathcal{T}$ also known as the domain shift problem. The SFCDCL is modeled as the class-incremental learning problem where each task carries disjoint classes $\mathcal{Y}_{t}^{s,tr}\cap\mathcal{Y}_{t'}^{s,tr}=\emptyset$ while no task identifiers are offered during the testing process. To comply to the issue of privacy, the access of labelled source-domain samples $\mathcal{D}_{t}^{s}$ is completely prohibited for domain adaptations. Only the pre-trained source model $f_{\psi}(g_{\theta}(.))$ and the unlabelled target domain samples $\mathcal{D}_{t}^{tr}$ are given for domain adaptations. $g_{\theta}(.):\mathcal{X}\rightarrow \mathcal{Z}$ is a feature extractor mapping the input space to the latent space, while $f_{\psi}(.): \mathcal{Z}\rightarrow \mathcal{Y}$ is a classifier converting the latent features to the output features.  

This learning problem opens at least three challenges: over-dependence on the pre-trained source model, noisy pseudo-labels and double catastrophic forgetting. The over-dependence on the pre-trained source model leads to sub-optimal solutions and also exist in the current source free domain adaptation literature. The noisy pseudo labels are seen because the learning process is only guided by unlabelled samples of the target domain and there exists the domain shift problem between the source domain and the target domain. The double catastrophic forgetting problem occurs because the pre-trained source model is continuously updated to keep pace with the learning sequences of the source domain in addition to the streaming tasks of the target domain. These challenges call for learning strategies to handle them concurrently.

\begin{figure*}
    \centering
    \includegraphics[width=0.9\linewidth]{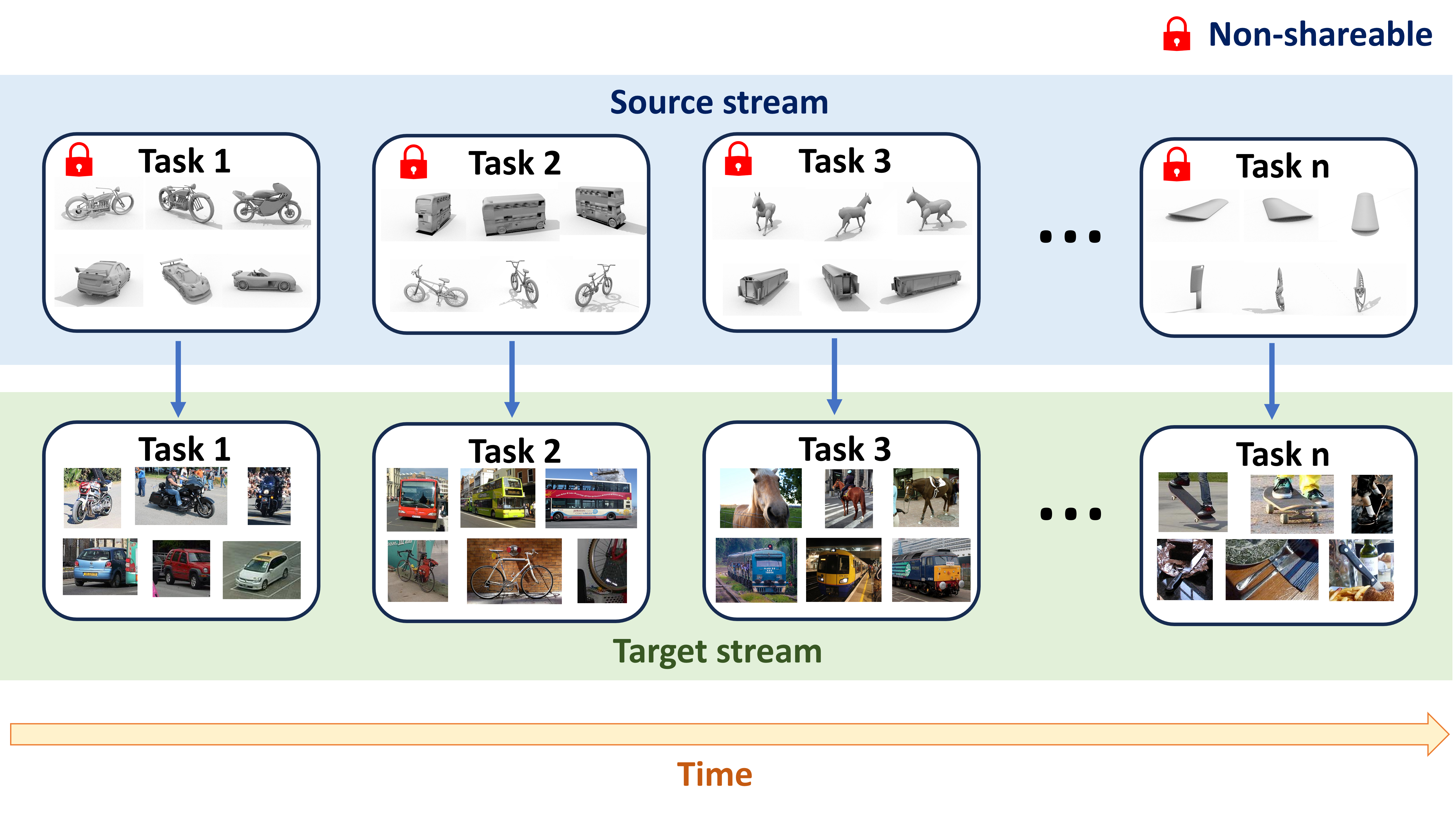}
    \caption{Source-Free Cross-Domain Continual Learning Problems: a model is presented with sequences of labeled learning tasks in the source domain and unlabeled learning tasks in the target domain where there exists the domain shift problem between the source domain and the target domain. A model is tasked to generalize well to the unlabeled learning task of the target domain without using any labeled learning tasks of the source domain for domain adaptations.}
    \label{fig:sfcdcl}
\end{figure*}

\subsection{Vision Language Model (VLM)}
REFEREE utilizes a VLM to overcome the over-dependence on the source model where CLIP \cite{Radford2021LearningTV} is applied. CLIP is structured by two encoders: visual encoder $g_{\theta_{v}}(.)$ and textual encoder $g_{\theta_{tx}(.)}$. The vision encoder $g_{\theta_{v}}(x)\in\Re^{d}$ generates the visual embedding, while the textual encoder generates the textual embedding $g_{\theta_{tx}(\omega_k)}\in\Re^{d}$ of a hand-crafted prompt of each class $\omega_k$, e.g., "A Photo of [CLASS]". The output of CLIP in term of classification probability is resulted from a matching degree of the visual embedding and the textual embedding. 
\begin{equation}\label{VLM}
    p_{VLM}(y|x)= \frac{\exp{(sim(g_{\theta_{v}}(x),g_{\theta_{tx}}(\omega_{k}))/\tau)}}{\sum_{k=1}^{M}\exp{(sim(g_{\theta_{v}}(x),g_{\theta_{tx}}(\omega_{k})))}}
\end{equation}
where $sim(.)$ is a similarity function, e.g., cosine similarity and $\tau$ is a temperature. CLIP is pretrained with millions of image-text pairs and ready to address the downstream tasks directly. Here, it functions as a zero-shot learner predicting the pseudo labels of target-domain samples. %We follow CoOp \cite{Zhou2021LearningTP} where a set of learnable prompts $\{v_i\}_{i=1}^{T}$ are inserted to the text encoder enhancing its generalization performance. That is, the input of the textual encoder is $V_{i}=[v_1, v_2, ..., v_{T},\omega_{k}]$. 

\subsection{Kernel Linear Discriminant Analysis (KLDA)}
The use of foundation model for CL has 
achieved prominent performances \cite{Zhou2024ContinualLW} because it generates generalized features improving predictive accuracy. Typically, the backbone network is frozen to safeguard from the CF problem and the adaptation to downstream task is availed by the application of fine-tuning techniques \cite{Wang2021LearningTP, Wang2022DualPromptCP} or even a simple nearest mean classification (NCM) scheme. Surprisingly, the NCM method using generalized features of the foundation model, i.e., the backbone network is frozen, surpasses complex fine-tuning approaches \cite{Zhou2024ContinualLW}. This finding motivates the development of KLDA \cite{Momeni2024ContinualLU} incorporating high-order statistics compared to the NCM approach only based on the prototypes. Using LDA directly over the generalized features is suboptimal because it creates linear decision boundary. This issue is tackled by the kernel method mapping the input data into a high-dimensional space where a data turns to be linearly separable. That is, the linear decision boundary in such space corresponds to a non-linear decision boundary in the original space. 

The mapping $\phi:\mathcal{Z}\rightarrow\mathcal{V}$, i.e., the generalized features are mapped to a high-dimensional space, where $\mathcal{V}$ is a potentially infinite-dimensional space, is defined via a kernel function $K(.)$ as follows:
\begin{equation}
    K(x_i,x_j)=(\phi(x_i),\phi(x_j))_{V}
\end{equation}
Suppose that the radial basis function (RBF) kernel is chosen, the kernel function is written as follows:
\begin{equation}
    K(x_i,x_j)=\exp{(-\frac{||x_i-x_j||^2}{2\sigma^2})}
\end{equation}
Note that the RBF kernel represents an inner product in an infinite-dimensional space. Nevertheless, this kernel trick requires the construction of the diagonal matrix which is intractable for the CL problem. This bottleneck can be addressed by approximating the mapping via the Random Fourier Features (RFF) method.

The RFF offers an efficient avenue to estimate the kernel function via the Bochner's theorem \cite{Momeni2024ContinualLU} where any continous, shift-invariant kernel can be formed by the Fourier transform. 
\begin{equation}
\begin{split}\label{fourier}
    K(x_i,x_j)=\int p(\omega)\exp{(i\omega^{\top}(x_i-x_j))}d\omega\\=\mathbb{E}_{\omega}[\exp{i\omega^{\top}(x_i-x_j)}]
\end{split}
\end{equation}
where $\omega,p(\omega)$ stand for the frequency in the Fourier domain and its corresponding probability density function respectively. \eqref{fourier} can be simplified using the Euler's formula given the fact that the kernel and the probability density function are real. 
\begin{equation}
    \mathbb{E}[z_{\omega}(x_i)z_{\omega}(x_j)]=K(x_i,x_j)
\end{equation}
\begin{equation}
    z_{\omega}(x)=\sqrt{2}\cos{(\omega^{\top}x+\beta)}
\end{equation}
where $\omega\backsim p(\omega),\beta\backsim \mathcal{U}(0,2\pi)$. The Fourier transform $p(\omega)$ for the RBF kernel is a normal distribution. Given $D$ pairs of $\omega,\beta$, the expectation can be estimated leading to the kerneled features. 
\begin{equation}\label{RFF}
    z(x)=\sqrt{\frac{2}{D}}[\cos{(\omega_{1}^{\top}x+\beta_1)},...,\cos{(\omega_D^{\top}x+\beta_{D})}]
\end{equation}
\begin{equation}
    z(x_i)^{\top}z(x_j)\approx K(x_i,x_j)
\end{equation}
where $\omega\backsim\mathcal{N}(0,\sigma^{-2}\mathbf{I}),\beta\backsim\mathcal{U}(0,2\pi)$. This approximation avoids the construction of the kernel matrix and the generalized features can be directly converted into random features, thereby in turn allowing for the application of the LDA method. The statistics of the LDA method, namely class mean $\mu_m$ and shared covariance matrix $A$, can be calculated:
\begin{equation}
    \mu_{m}=\frac{1}{n_m}\sum_{i=1}^{n_m}Z_{i}
\end{equation}
\begin{equation}
    A=\frac{N_{prev}}{N_{total}}A+\frac{1}{N_{total}}\sum_{i=1}^{n_m}(Z_i-\mu_{m})(Z_{i}-\mu_{m})^{\top}
\end{equation}
where $N_{prev}=N_{total},N_{total}=N_{prev}+n_{m}$. $n_{m}$ denotes the number of samples of the $m-th$ class. Note that the covariance matrix $A$ is shared across all classes and calculated recursively. Once obtaining the statistics, the weights and bias of the classifier are enumerated afterward. 
\begin{equation}
    w_{m}=A^{-1}\mu_{m}
\end{equation}
\begin{equation}
    b_{m}=-\frac{1}{2}\mu_{m}^{\top}A^{-1}\mu_{m}
\end{equation}
The classification function is defined as follows:
\begin{equation}
    F(x)=z(x)^{\top}W+b
\end{equation}
where each element of $F(x)$ corresponds to the score of a class. The classification is done by selecting the class having the highest score. The KLDA is implemented here as a backbone network to cope with the double catastrophic forgetting problem, i.e., source and target. That is, it is applied in the source domain as one of the branch of the dual-branch network structure and the target domain to address the unlabeled target domain. Nevertheless, the naive application of KLDA produces suboptimal performances because of the noisy pseudo label problem and the over-dependence on the source pre-trained model. Hence, it is extended here by incorporating the Vision-Language Model (VLM) under the dual-branch network structure, the frequency-aware prompting technique and the uncertainty-aware weighting scheme to combat the noisy pseudo label problem due to the domain shift problem.  

\section{Methodologies}
\subsection{Overview}
REFEREE is built upon a dual-branch network structure to overcome an over-dependence on the source-pretrained model where its outputs are aggregated from the source-pretrained branch and the VLM branch. The source-pretrained branch is modeled by Vision Transformer (ViT) and meant to capture domain-specific characteristics while CLIP is implemented in the VLM branch and intended to extract domain-invariant characteristics. The source-pretrained branch is fully learnable to determine the mean and covariance matrix of the KLDA leaving its backbone network fixed while the VLM branch are frozen \cite{Zhou2021LearningTP}. Because of the problem of domain shifts, noisy pseudo labels are unavoidable. Such problem is attacked by the frequency-aware prompting technique. That is, an image is transformed into a frequency domain and decomposed into low-frequency and high-frequency components. The low-frequency component is kept unchanged because it represents stable information which can be safely learned, whereas the high-frequency component is modified by zeroing and randomly sampling because the high frequency component contains noisy information. 

\begin{figure*}
    \centering
    \includegraphics[width=0.9\linewidth]{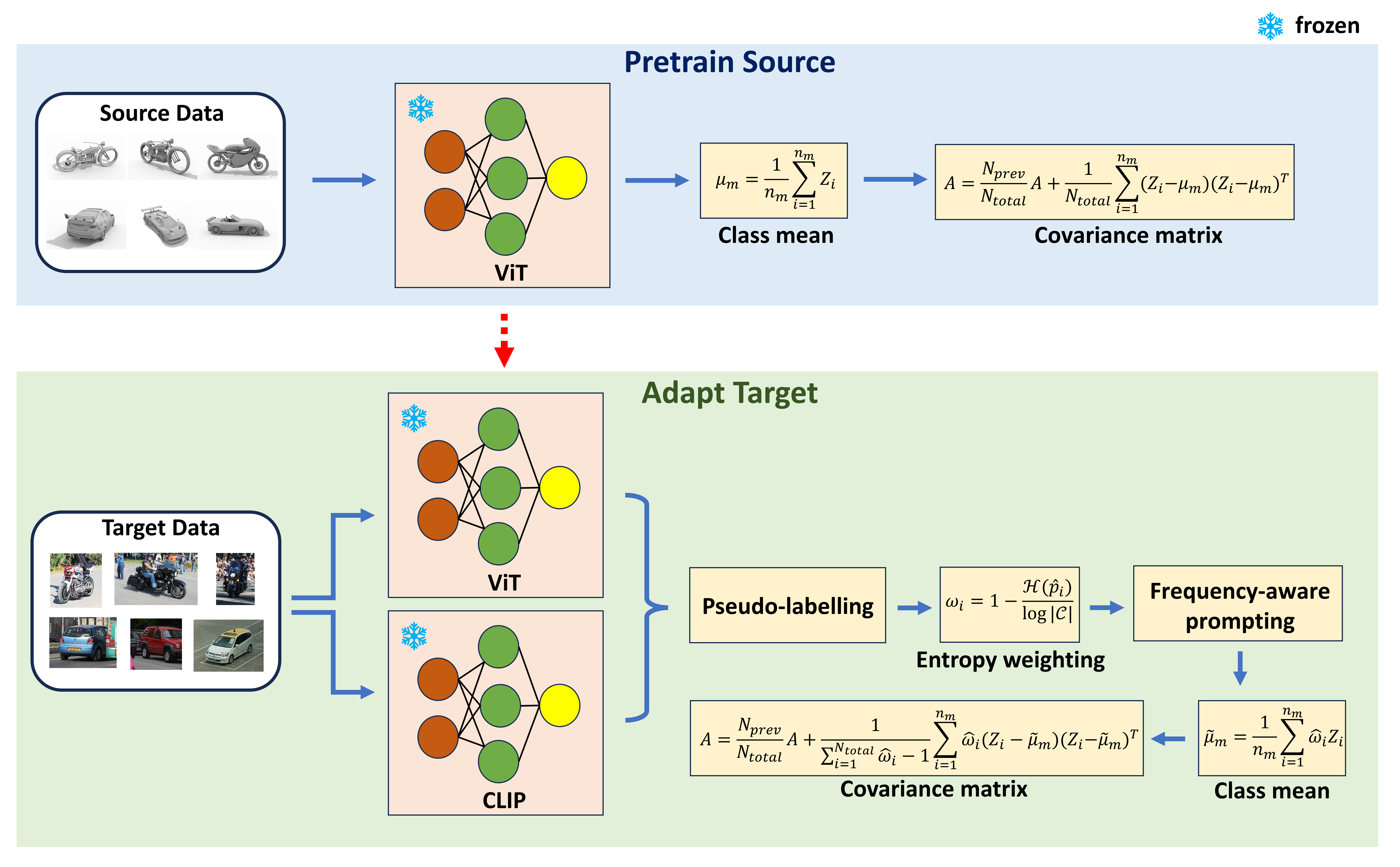}
    \caption{REFEREE architecture. A dual-branch framework couples a frozen source-pretrained encoder (ViT) with a frozen vision--language model (CLIP). For each \emph{target} image, both branches produce scores that are fused $\hat{p}=\alpha\,p+\beta\,p_{\text{VLM}}$ to obtain pseudo-labels and entropies. A frequency-aware prompting module (DWT) suppresses high-frequency components and augments low-frequency content. The fused confidence defines uncertainty weights $\omega\!\in\![0,1]$ that weight the class means $\mu$ and shared covariance $\Sigma$ for KLDA over random Fourier features; these statistics update the target classifier across tasks while the backbones remain frozen. No source-domain samples are used during target adaptation.}

    \label{fig:sfcdcl}
\end{figure*}

The noisy pseudo-label problem is further overcome with the uncertainty-aware weighting strategy. That is, the uncertainty weights are derived by using the Shannon entropy technique and applied to weight the mean and covariance matrix of the KLDA approach to alleviate the noisy pseudo label problem. Last but not least, the double catastrophic forgetting problem is addressed using the idea of kernel linear discriminant analysis (KLDA) \cite{Momeni2024ContinualLU}. That is, we rely on generalized features of the foundation model frozen during the training process to combat the DCF problem. The original LDA results in suboptimal performances because it constructs the linear decision boundary. This problem can be solved using the kernel method mapping the feature into a higher-dimensional space where the data becomes linearly separable. Nonetheless, the calculation of the kernel matrix is intractable in the continual learning setting. The mapping can be estimated using Random Fourier Features (RFF) \cite{Momeni2024ContinualLU}. Hence, the class mean and the shared covariance matrix are built upon the random features leading to the weights and biases of the LDA. Such trick is applied for both source and target domain. Note that no source samples are provided while learning the target domain. In other words, the construction of KLDA parameters are done independently, i.e., the source KLDA is applied to induce the pseudo-labels of the target domain and drives the ViT branch. Ours is gradient-free where the training process is localized to calculate the mean and covariance matrix.  %the double catastrophic forgetting problem is addressed using the notion of statistical sampling.  That is, only statistics of the learning tasks are stored in the memory rather than raw images, thus significantly alleviating memory burdens. The experience replay mechanism is induced by random sampling of the statistics. Since the model is continually adjusted, it paves a possibility of outdated statistics. A projector is implemented to map the old statistics to a new feature space followed by the refinements of the old statistics.    

\subsection{Dual Branch Network Structure}
REFEREE is structured as a dual-branch network structure comprising the source-pretrained branch and the VLM branch \cite{Zhan2024TowardsDC}. The source-pretrained branch $f_{\psi}(g_{\theta}(.)): \mathcal{X}\rightarrow\mathcal{Y}$ converts the input image to the output logits, while the VLM branch mines two different modalities into a joint embedding by means of the vision encoder $g_{\theta_{v}}(.):\mathcal{X}\rightarrow\mathcal{Z}$ and the textual encoder $g_{\theta_{tx}}(.):\mathcal{V}\rightarrow\mathcal{Z}$. That is, the vision encoder receives an input image while the textual encoder processes the hand-crafted prompts of each class $\omega_k$. The final prediction is drawn from the weighted mixture of the two network outputs \cite{Wang2023FewShotCS}. 
\begin{equation}\label{pred}
    \hat{p}_{i}=\alpha p(y_i=c|x_i) + \beta p_{VLM}(y_i=c|x_{i})
\end{equation}
where $p(y_i=c|x_i)=\sigma(f_{\psi}(g_{\theta}(x_i)))$ and $\sigma(.)$ is a softmax function, while $p_{VLM}(y_i=c|x_i)$ is obtained from (\ref{VLM}). $\alpha=\frac{\arg\max_{k\in[1,M]}\alpha_k}{\arg\max_{k\in[1,M]}\alpha_k+\arg\max_{k\in[1,M]}\beta_k}$ and  $\beta=\frac{\arg\max_{k\in[1,M]}\beta_k}{\arg\max_{k\in[1,M]}\alpha_c+\arg\max_{k\in[1,M]}\beta_k}$. $\alpha,\beta$ respectively stand for the soft labels produced by $p,p_{VLM}$. This strategy enables both domain-invariant and domain-specific information to be taken into account when inducing the final predictions. On the other sides, the parameters of CLIP are frozen, while the source-pretrained branch is fully learned only to determine the KLDA parameters leaving the backbone network fixed. 

\subsection{Frequency-aware Prompting Technique}
Inspired by \cite{Zhang2024ExploringCF}, the frequency-aware prompting method is developed to combat the noisy pseudo label problem. This method encourages low-frequency components while discouraging high-frequency components, which are vulnerable to noisy information. That is, the high-frequency components are converted either with normal distribution sampling or with zeros, thus generating augmented samples. In other words, this learning scheme induces a new learning task improving learning diversities. Note that pseudo labelled samples are usually less than the original task size.

Given a pseudo-labelled learning task of the target domain $\mathcal{D}_{t}^{tr}=\{(x_i,\hat{y}_{i})\}_{i=1}^{N_{t}^{tr}}$ where $\hat{y}_{i}$ is a pseudo label of the $i-th$ image induced by applying a threshold to the model's prediction (\ref{pred}), the discrete wavelet transform (DWT) is applied to the image $x_i$. This mechanism leads to four frequency sub-bands $f_{LL},f_{LH},f_{HL},f_{HH}$. The low-frequency component $f_{ll}$ is unchanged while $f_{high}=[f_{LH},f_{HL},f_{HH}]$ are replaced with the same size of tensor of zeros and with the same size of tensor of random numbers sampled from normal distributions. Thereafter, the inverse discrete wavelet transform strategy is applied to produce reconstructed images $x_{i}^{zeros}$ and $x_{i}^{rand}$. an augmented sample is generated as $\tilde{x}_{i}=\{x_{i},x_{i}^{zeros},x_{i}^{rand}\}$ where they share the same label $\hat{y}_{i}$. 

It is evident that $\tilde{X}$ contains diverse information and less noisy than the original samples $X$ consisting of noisy samples near the decision boundary. This strategy does not introduce semantic drift because the low frequency component is unchanged. Rather, it promotes low frequency components for the learning process, thereby making possible for robust inductive bias. These samples are further refined over $T_{max}$ iterations using the gradient ascent procedure as the adversarial learning step \cite{Wang2021CrossDomainFC} to generate challenging tasks. 
\begin{equation}\label{wavelet}
    \tilde{\mathcal{D}}_{t}^{tr}=\{X_{T_{max}},X_{T_{max}}^{zeros},X_{T_{max}}^{rand},Y\}
\end{equation}
where $\tilde{\mathcal{D}}_{t}^{tr}$ stands for a challenging version of the $t-th$ learning task of the target domain. Moreover, the mutual attention strategy is applied \cite{Zhan2024TowardsDC} where it stresses the correlated features while suppressing other unrelated information. That is, the low frequency components are preserved and stressed whereas other uncorrelated frequency features are suppressed.  

\subsection{Uncertainty-aware Weighting Scheme}
Because of the domain shift problem, the generation of pseudo labels as per \eqref{pred} contains noises leading to miscalculation of the means and covariance matrix of the KLDA, although our approach is gradient-free. To remedy this situation, the uncertainty-aware weighting scheme is proposed based on the \emph{normalized Shannon entropy} of the fused prediction, following the spirit of entropy normalization used in \cite{vu2019advent}. This strategy reduces the effect of noisy pseudo labels when the uncertainty is high by allocating low weights in the calculations of the mean and covariance matrix, whereas a high weight is assigned if the uncertainty is low. The uncertainty weight $\omega_{m}\in[0,1]$ is applied as follows:
\begin{equation}\label{weighted_mean}
    \mu_{m}=\frac{1}{n_{m}}\sum_{i=1}^{n_m}\hat{\omega}_{i}Z_{i}
\end{equation}
\begin{equation}\label{weighted_covariance}
   A=\frac{N_{prev}}{N_{total}}A+\frac{1}{\sum_{i=1}^{N_{total}}\hat{\omega}_{i}-1}\sum_{i=1}^{n_{m}}\hat{\omega}_{i}(Z_{i}-\mu_{m})(Z_{i}-\mu_{m})^{\top} 
\end{equation}
 $\hat{\omega_{i}}$ is determined from the \emph{linear, normalized entropy weight} in $[0,1]$. $N_{prev}=N_{total};N_{total}=N_{prev}+n_{m}$. Suppose that $\mathcal{H}(.)$ is the Shannon uncertainty function, the uncertainty weight is derived as follows:
 % \begin{equation}
 %     \omega_{i}=\frac{1}{\mathcal{H}(\hat{p}_{i})}
 % \end{equation}
 \begin{equation}\label{w}
    \omega_i \;=\; 1 \;-\; \frac{\mathcal{H}(\hat{p}_{i})}{\log |\mathcal{C}|}.
\end{equation}
 where $\mathcal{C}$ is current task’s class set, $\hat{p}_{i}$ is a prediction confidence computed as per \eqref{pred}. The higher the entropy implies the less confidence the prediction is whereas the lower the entropy means the more confidence the prediction is. It is intuitive to assign a high weight when the uncertainty is low. In addition, this setting is also less prone to numerical instability than the inverse entropy because of no risk of division with zero. 
%  The uncertainty weight is further normalized to assure its correct range. 
% \begin{equation}
%     \hat{\omega}_{i}=\frac{\omega_{i}}{\max_{i=1,...,N_{total}}\omega_{i}}
% \end{equation}

 Because REFEREE is gradient-free where only the mean and covariance matrix are enumerated during the training process, the uncertainty weight is fixed once determined, i.e., no training epoch is implemented in our approach.

\begin{algorithm}
\caption{REFEREE}
\label{alg:referee-target}
\begin{algorithmic}
\REQUIRE Labeled source tasks $\{\mathcal{D}^s_t\}_{t=1}^{T}$ with class sets $\mathcal{C}^s_t$; Unlabeled target tasks $\{\mathcal{D}^{tr}_t\}_{t=1}^{T}$ with class sets $\mathcal{C}^{tr}_t$; frozen $f_{\psi}, g_{\theta_{v}}, g_{\theta_{tx}}$; RFF params $(D,\sigma)$; prompts $\{\omega_c\}$ and text feats $E=[g_{\theta_{tx}}(\omega_c)]$; Source KLDA $(\mu_m,A)$; shared RFF $(\omega,b)$; wavelet $W$
\ENSURE Target KLDA $(\tilde{\mu}_m, \tilde{A})$
\STATE \textbf{Pretrain Source}
\STATE Initialize $n_m \gets 0$, $\mu_m \gets 0\in\mathbb{R}^D$ for all classes; $A \gets 0\in\mathbb{R}^{D\times D}$
\STATE Sample RFF $(\omega,b)$ for RBF 
% define $\phi(z)=\sqrt{2/D}\cos(z\omega+b)$
\FOR{$t=1$ to $T$}
  \FORALL{mini-batch $(x,y)\subset \mathcal{D}^s_t$}
    % \STATE $z \gets \mathrm{normalize}(f(x))$;\quad $\Phi \gets \phi(z)$
    \STATE $z \gets RFF(g_{\theta}(x)) $
    \STATE Calculate mean ($\mu_m$) $\gets$ eq.(9) and covariance matrix ($A$) $\gets$ eq.(10)
  \ENDFOR
\STATE \textbf{Adapt Target}
  \STATE $\mathcal{C} \gets \mathcal{C}^{tr}_t$;\quad initialize $\tilde{n}_m\!\gets\!0$, $\tilde{\mu}_m\!\gets\!0$, $\tilde{A}\!\gets\!0$
  \FORALL{mini-batch $x \subset \mathcal{D}^{tr}_t$}
    \STATE \textbf{(A) Pseudo-labels \& uncertainties (unsupervised)}
    % \STATE $z \gets \mathrm{normalize}(f(x))$;
    % \STATE $\Phi \gets \phi(z)$
    \STATE \emph{Source-pretrained branch:} $z \gets RFF(g_{\theta}(x))$; compute $p$
    \STATE \emph{VLM branch:}$z_v \gets g_{\theta_{v}}(x)$; $z_{tx} \gets g_{\theta_{tx}}(\omega_c)$; compute $p_{VLM} \gets$ eq.(1)
    \STATE \emph{Per-sample fusion:} calculate $\alpha,\beta$
    \STATE Calculate $\hat{p} \gets$ eq.(14)
    % $p^{\mathrm{fuse}}=\alpha\,p^{\mathrm{klda}}+\beta\,p^{\mathrm{clip}}$
    \STATE \textbf{(B) Weighted KLDA updates}
    \STATE $\tilde{x}\gets \mathrm{DWT}(x;W)$ \COMMENT{frequency aware prompting}
    % \STATE $z' \gets \mathrm{normalize}(f(\tilde{x}))$;\quad $\Phi' \gets \phi(z')$
    \STATE Calculate weight $w \gets$ eq.(18)
    \STATE Apply weight to the mean ($\tilde{\mu}_m$) and covariance matrix ($\tilde{A}$)
  \ENDFOR
\ENDFOR
\end{algorithmic}
\end{algorithm}

\subsection{Algorithm}
An overview of REFEREE's learning policy is described in the algorithm \ref{alg:referee-target}. Because of the presence of two CL processes in the source and target domains, there exists two training loops for the source and target domains. Note that the source-free cross-domain continual learning features different but related CL processes where no label is offered in the target domain and no source domain samples are shared for domain adaptations. This facet implies the applications of two KLDAs for the source domain and the target domain. 

The learning process begins with the CL process of the source domain where labeled samples exist. It follows the standard protocol of KLDA where at first random features $Z$ are generated by transforming the original generalized features via the Fourier transform \eqref{RFF}. It is followed by calculation of the mean and covariance matrix leading to the weights and biases of the LDA while leaving the backbone network fixed. Once done, the pseudo labels are generated by a mixture of the KLDA's output of the source domain and the zero-shot learner, i.e., CLIP as per \eqref{pred}.

After that, pseudo-labeled samples undergo the Wavelet decomposition where the low frequency components are promoted while suppressing the high frequency components \eqref{wavelet} to overcome the noisy pseudo-label problem due to the domain shift issue. As with the source domain, the unlabeled target domain is handled by applying the random Fourier transformation to the generalized features. To further reduce the impacts of noisy pseudo labels, the uncertainty-aware weighting scheme is implemented where the uncertainty weight is enumerated as per \eqref{w} and in turn applied to the calculations of the mean and covariance matrix as per \eqref{weighted_mean} and \eqref{weighted_covariance}.   

\subsection{Complexity Analysis}
Let $N$ be the total number of target samples, split into $B$ mini-batches with $\sum_{b=1}^{B} N_b = N$.
Let $D$ be the random Fourier characteristic (RFF) dimension, $k$ be the number of augmentations, $H$, $W$ is the height and width of the image (in pixels), respectively. 
REFEREE performs a single pass over the target data (gradient-free). Following
pseudocode in Algorithm 1, the REFEREE method consists of several processes, per batch it performs:
(i) RFF mapping, (ii) pseudo-label fusion \& weighting, (iii) KLDA mean/covariance updates, and (iv) frequency-aware prompting (FAP). Denote by $C(\cdot)$ the cost of each process in one batch.

\begin{equation}
\begin{aligned}
C(\text{REFEREE})
&= \sum_{b=1}^{B}\Big(C(\text{RFF}) + C(\text{Fusion+Weight})\\ 
&\hspace{1.25em}+ C(\text{KLDA mean})
+ C(\text{KLDA covariance})\\ 
&\hspace{1.25em}+ C(\text{FAP})\Big)\\[2pt]
&= \sum_{b=1}^{B} N_b \Big(\mathcal{O}(D) + \mathcal{O}(|\mathcal{C}_t|) + \mathcal{O}(D)
+ \mathcal{O}(D^2) \\
&\hspace{1.25em}+ \mathcal{O}(kHW)\Big).
\end{aligned}
\end{equation}

\begin{equation}
\begin{aligned}
C(\text{REFEREE})
&= \mathcal{O}\!\Big(D \sum_{b=1}^{B} N_b\Big)
+ \mathcal{O}\!\Big(|\mathcal{C}_t| \sum_{b=1}^{B} N_b\Big)\\
&\hspace{1.25em}+ \mathcal{O}\!\Big(D^2 \sum_{b=1}^{B} N_b\Big)+ \mathcal{O}\!\Big(kHW \sum_{b=1}^{B} N_b\Big).
\end{aligned}
\end{equation}

Substituting $\sum_b N_b = N$:

\begin{equation}
\begin{aligned}
C(\text{REFEREE})
&= \mathcal{O}(N D^2)
+ \mathcal{O}(N D)
+ \mathcal{O}(N |\mathcal{C}_t|)\\
&\hspace{1.25em}+ \mathcal{O}(N kHW).
\end{aligned}
\end{equation}

In practice, $|\mathcal{C}_t|$, $k$, $H$, $W$, and $D$ are fixed (constants with respect to $N$). Therefore, REFEREE’s runtime scales \emph{linearly} with the
number of target samples in the single pass:
\[
C(\text{REFEREE}) \approx \mathcal{O}(N).
\]

\begin{figure*}
    \centering
    \includegraphics[width=0.9\linewidth]{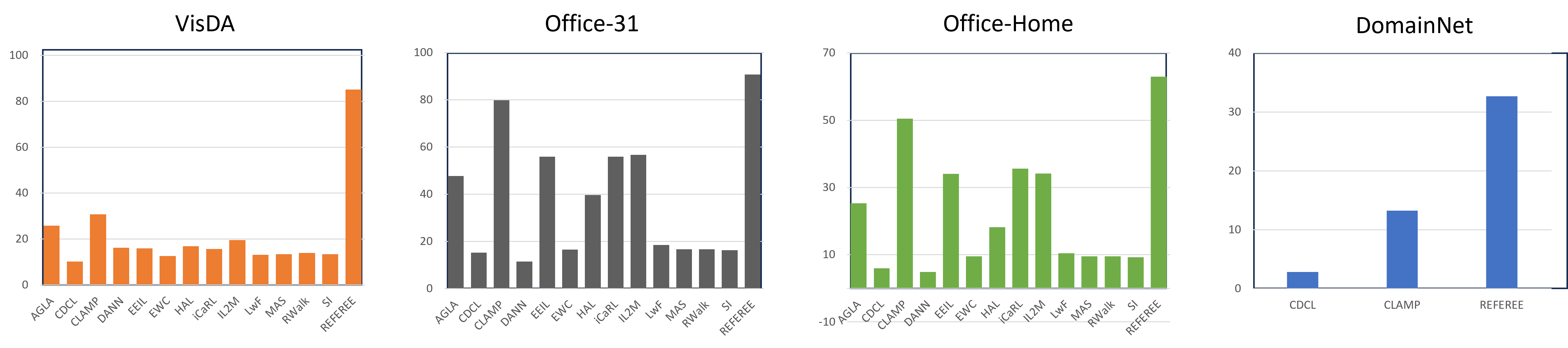}
    \caption{Average accuracy (\%) comparison across different baselines for four datasets: VisDA, Office-31, Office-Home, and DomainNet}
    \label{fig:bar}
\end{figure*}

\section{Experiments}
This section discusses our numerical study encompassing datasets, comparisons with published works, numerical results, ablation study, and sensitivity analysis.

\begin{table}[t]
\centering
\caption{Dataset descriptions.}
\begin{tabular}{c|c|c|c}
\hline
Dataset     & \# classes & \# tasks & \# classes per task \\ \hline
VisDA       & 12         & 4        & 3                   \\
Office-31   & 31         & 5        & 6                   \\
Office-Home & 65         & 13       & 5                   \\
Domain-Net  & 345        & 15       & 23                  \\ \hline
\end{tabular}
\label{tab:dataset}
\end{table}

\subsection{Datasets}
Our algorithms is tested across 21 cross-domain continual learning problems: 1) the Office-31 problem \cite{Saenko2010AdaptingVC} comprises three domains, Amazon (A), DSLR (D), Webcam (W) and 4652 colored images. Each domain possesses 31 classes and is divided into 5 tasks with 6 classes each except for the last task with 7 classes; 2) the Office-home problem \cite{Venkateswara2017DeepHN} presents four domains, Artistic image (Ar), Clip art (CI), Product images (Pr) and Real-World Images (RW) where each domain has 65 classes and is split into 13 tasks, i.e., each task carries 5 classes; 3) the VisDA 2017 problem \cite{Peng2017VisDATV} presents a source domain of the synthetic 2D rendering of 3D models generated from different angles and lighting conditions while a target domain features a photo-realistic image. This problem has 12 classes where each task carries 3 classes of 4 tasks; 4) the DomainNet problem \cite{DomainNet} is a large-scale problem with 345 classes and 6 domains, Clipart (clp), Infographics (inf), Painting (pnt), Quickdraw (qdr), Real (rel), and Sketch (skt). Because of limited computational resources, only two domains are tested in this paper, clipart$\leftrightarrow$sketch. This problem is formulated into 15 tasks of 23 classes each. Characteristics of these datasets are summed up in Table \ref{tab:dataset}.

\subsection{Benchmark Algorithms}
Our algorithm, REFEREE, is compared against 13 prior arts: CLAMP \cite{Weng2024CrossDomainCL} (Information Sciences 2024), CDCL \cite{VinciusdeCarvalho2024TowardsCC} (ICDE 2024), LWF \cite{Li2016LearningWF} (IEEE TPAMI 2016), SI (ICML 2017) \cite{Zenke2017ContinualLT}, MAS \cite{aljundi2018memory} (ECCV 2017), RWalk \cite{Chaudhry2018RiemannianWF} (ECCV 2018), HAL \cite{Chaudhry2019UsingHT} (AAAI 2021), AGLA \cite{Masum2023AssessorGuidedLF} (Information Sciences 2023), ICARL \cite{Rebuffi2016iCaRLIC} (CVPR 2016), DANN \cite{Ganin2015DomainAdversarialTO} (JMLR 2015), IL2M \cite{Belouadah2019IL2MCI} (ICCV 2019), EEIL \cite{Castro2018EndtoEndIL} (ECCV 2018), EWC \cite{Kirkpatrick2016OvercomingCF} (PNAS 2016). Comparisons are done under the same computational environments, i.e., NVIDIA A5000 GPU with 24 GB of RAM, where all algorithms are independently evaluated across five random seeds. The reported numerical results are averaged from the five independent runs. The hyper-parameters of consolidated algorithms are selected using the grid search approach to ensure fair comparisons.

% \begin{table*}[!t]
% \caption{Average accuracy (\%) of all learned tasks on DomainNet dataset}\label{tab:dom}
% \centering
% \begin{tabular}{c|c|c}
% \hline
% Method           & clipart→sketch & sketch→clipart \\ \hline
% CDCL             & 3.49           & 2.17           \\
% CLAMP            & 12.87          & 13.67          \\ \hline
% \textbf{REFEREE} & \textbf{24.98} & \textbf{31.05} \\ \hline
% \end{tabular}
% \end{table*}

\subsection{Evaluation Metrics}
The performance of consolidated algorithms is assessed using the average accuracy of the unlabeled target domain to analyze model performances in facing the domain shift problem and the double catastrophic forgetting problem. Note that all algorithms except our approach, REFEREE, utilizes the source domain samples during the domain adaptation step whereas REFEREE is a completely source-free approach where no source samples are exploited when performing the domain adaptation step. The average accuracy $\mathcal{A}_{cc}\in[0,1]$ is written as follows:
\begin{equation}
    \mathcal{A}_{cc}=\frac{1}{k}\sum_{j=1}^{k}a_{k,B_{k,j}}
\end{equation}
where $a_{k,B_{k,j}}\in[0,1]$ stands for the accuracy induced by the test set of task $j$ after the model is trained with the task $k$. That is, $\mathcal{A}_{T}$ concerns on the accuracy of all tasks. 
\subsection{Implementation Details}
REFEREE is constructed under a backbone network of ViT-B (12 layers, 768 dimensions) and CLIP whereas other algorithms utilize ResNet34 for Office-31 and VisDA; and ResNet50 for Office-Home and DomainNet. We apply a pretrained model of ImageNet to ensure its generalization power for all consolidated algorithms. Besides, the two hyper-parameters of REFEREE, $D,\sigma$, are set to $D=6000$ and $\sigma=10^{-4}$ for all experiments in this paper.  

\begin{figure}
    \centering
    \includegraphics[width=0.9\linewidth]{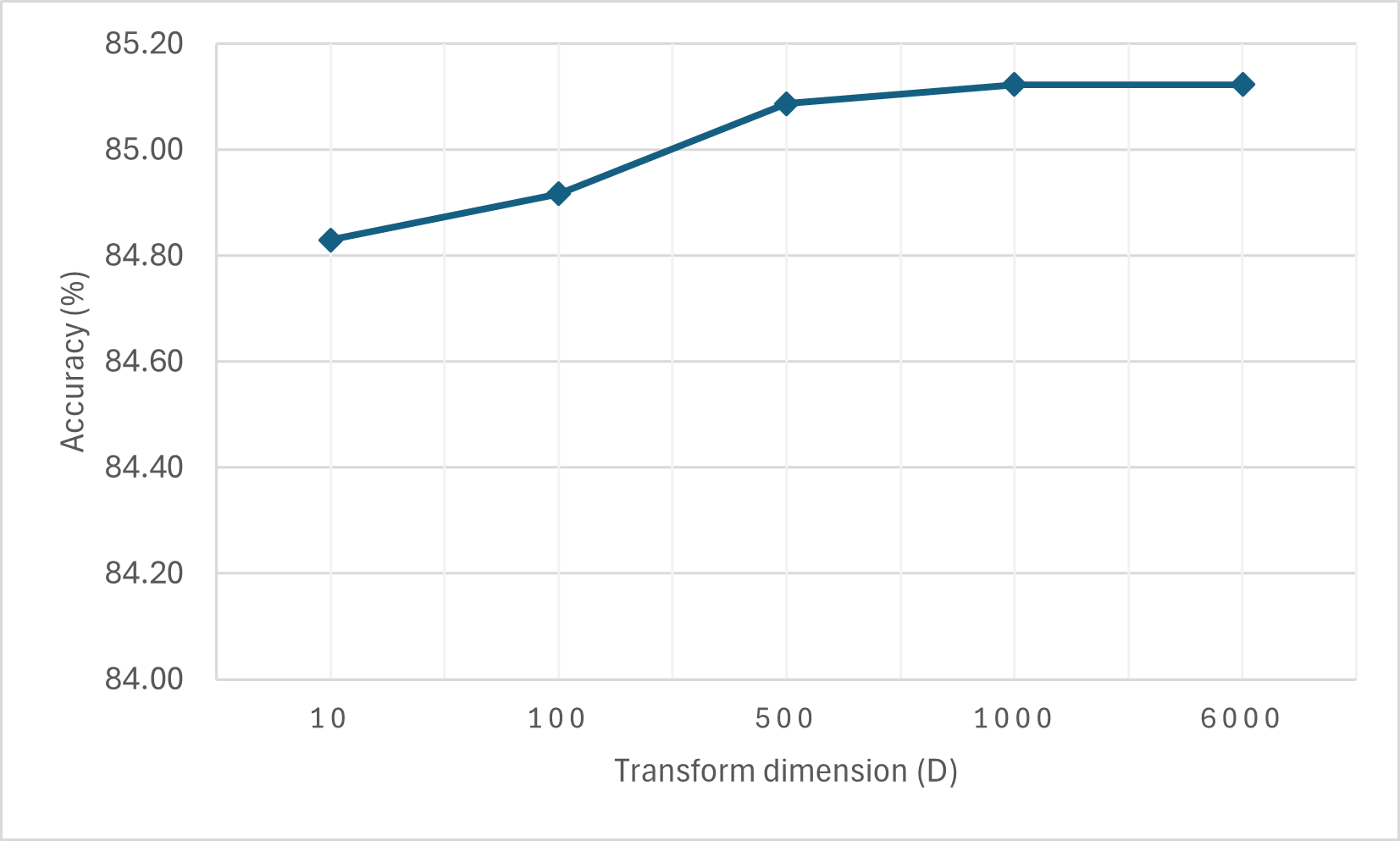}
    \caption{Sensitivity to transform dimension $D$. Performance saturates from $D\!\approx\!500$ onward.}

    \label{fig:sens-D}
\end{figure}

\begin{figure}
    \centering
    \includegraphics[width=0.9\linewidth]{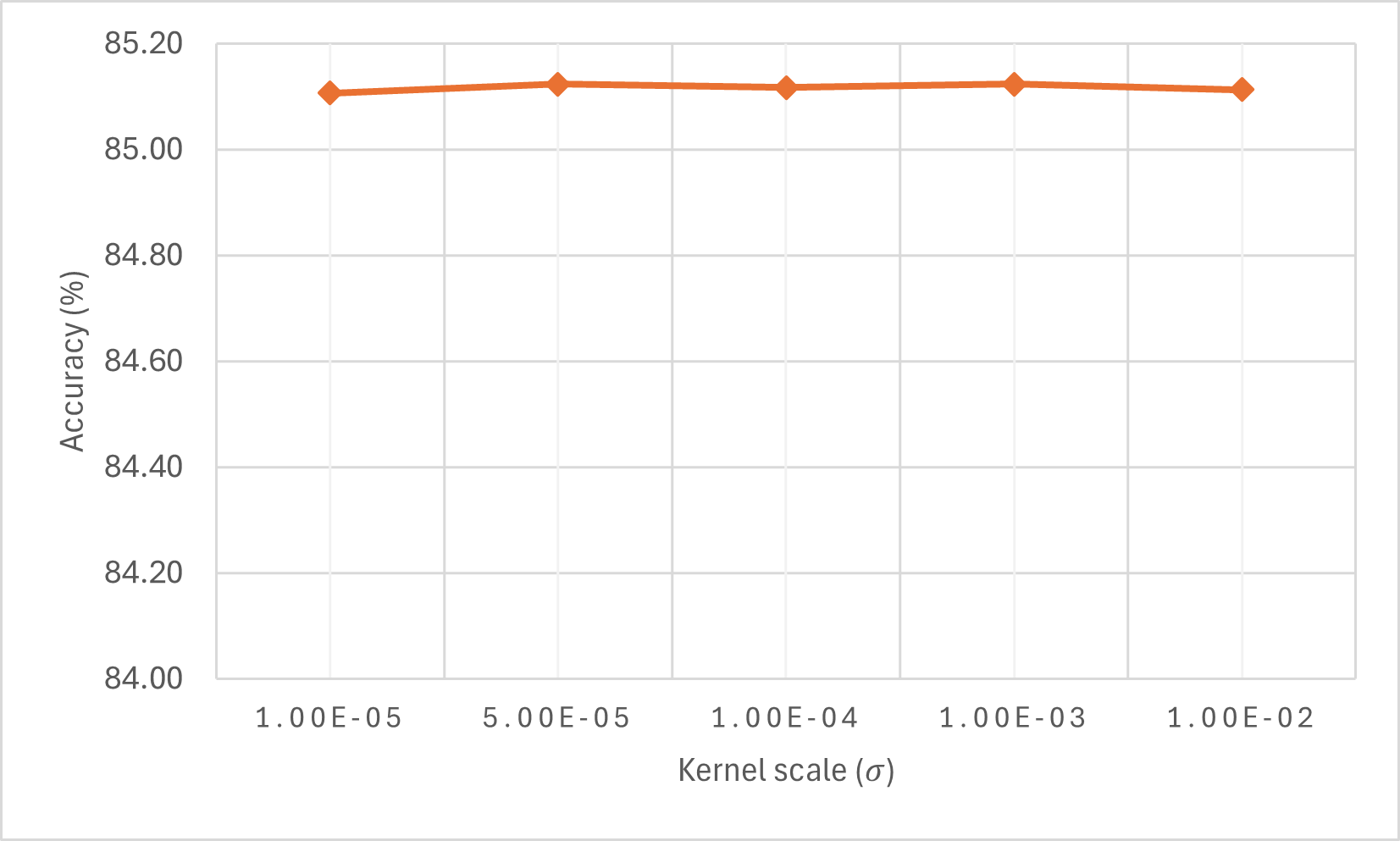}
    \caption{Sensitivity to kernel width $\sigma$ in KLDA. Accuracy is flat across four orders of magnitude.}

    \label{fig:sens-Sigma}
\end{figure}

\subsection{Numerical Results}
The advantage of our algorithm, REFEREE, is clearly demonstrated where it beats prior arts with significant margins across all 21 problems. In the VisDA problem, REFEREE exceeds CLAMP in the second place with over $54\%$ margin as presented in Table \ref{tab:dig}. Note that CLAMP enjoys the access of source-domain samples whereas REFEREE is a completely source-free approach. Close to the $11\%$ margin is observed in the office31 problem as presented in Table \ref{tab:off31} between REFEREE and CLAMP while over 12$\%$ gap exists between REFEREE and CLAMP as presented in Table \ref{tab:hom} for the office home problem. Last but not least, REFEREE beats CLAMP and CDCL with $19\%$ gap in the complex problem, the DomainNet problem. In a nutshell, the gap between REFEREE and prior arts becomes wider in the complex problem such as the DomainNet, VisDA and OfficeHome having more classes and tasks confirming the efficacy of our approach for source-free cross-domain continual learning problems. Fig. \ref{fig:bar} visualizes the numerical results of all consolidated algorithms across 4 datasets: VisDA, Office31, OfficeHome, DomainNet.  

\begin{table}[t]
\centering
\caption{Ablation on REFEREE components.}
\setlength{\tabcolsep}{10pt}
\begin{tabular}{lc}
\toprule
\textbf{Method} & \textbf{Avg.\ (\%)} \\
\midrule
\textbf{\textsc{Referee}} & \textbf{85.12 $\pm$ 0.4} \\
\textsc{Referee} (w/o CLIP) & 68.60 $\pm$ 0.4 \\
\textsc{Referee} (w/o augmentation) & 83.79 $\pm$ 0.7 \\
\textsc{Referee} (w/o weighting) & 84.44 $\pm$ 0.4 \\
\bottomrule
\end{tabular}

\label{tab:ablation}
\end{table}

\subsection{Ablation Study}
We quantify the contribution of each component in REFEREE: (i) the CLIP branch (vision–language scores), (ii) frequency-aware augmentation (FAP), and (iii) uncertainty-aware weighting. Our experiments are performed with the ViSDA dataset possessing challenging characteristics. Removing the CLIP branch causes the largest drop (over 16\% gap), confirming that language–image alignment substantially improves pseudo-label quality on target data. Disabling FAP yields a smaller but consistent decrease ($1.33 \%$ gap) and increases variability (std $0.7$), indicating FAP improves robustness across runs. Turning off uncertainty-aware weighting reduces performance by $0.68$\%, showing that entropy-based reweighting further mitigates noise in pseudo-labels. Overall, the full REFEREE
combines complementary gains: CLIP provides strong semantic priors, FAP enhances cross-domain invariance, and weighting stabilizes updates by down-weighting uncertain samples.

\subsection{Robustness Analysis}
This subsection concerns on the analysis of robustness against different hyper-parameters of REFEREE, namely $D,\sigma$ respectively denoting the dimension of random Fourier features (RFF) and the spread of the normal distribution of the Fourier frequency $\omega\backsim\mathcal{N}(0,\sigma^{2}\mathbf{I})$. Note that the two hyper-parameters are the only predefined parameters of REFEREE and our experiments are performed using the VisDA problem. It is seen that REFEREE is robust against variations of hyper-parameters where the accuracy is relatively stable at around $85\%$ as shown by Fig. \ref{fig:sens-D}, i.e., only $1\%$ reductions in accuracy occur when selecting $D$ to be lower than $<500$. On the other hand, REFEREE is not sensitive to selections of the spread of the normal distributions where differences of classification rates are subtle for different spreads of the normal distributions as exhibited by Fig. \ref{fig:sens-Sigma}. This finding confirms the robustness of REFEREE against different hyper-parameters. %For all our experiments, the two hyper-parameters are simply fixed at $D=6000, \sigma=10^{-4}$. 

\section{Conclusion}
This paper proposes a source-free cross-domain continual learning problem including its solution, REFEREE. REFEREE is built upon a dual-branch network structure where three exist a ViT branch and a VLM branch benefiting from a zero-shot learning aptitude of CLIP. The frequncy-aware prompting technique is implemented to address the domain-shift problem where an image is decomposed into low and high frequency components. Low frequency components are promoted whereas high-frequency components are suppressed. The uncertainty-aware weighting technique is applied to further alleviate the detrimental effect of noisy pseudo labels where the uncertainty weights are derived from the Shannon entropy concept and introduced to weight the mean and covariance matrix of KLDA. Our approach is gradient-free where the backbone network is kept frozen while benefiting from the random Fourier features (RFF) to generate the mean and covariance matrix. In addition, our approach constitutes a fully source-free approach and is applicable in the black-box scenario. Rigorous experiments have demonstrated the advantage of our approach where it beats recently published algorithms with significant margins particularly in the complex problems: VisDA, OfficeHome and DomainNet. Our future work is devoted to study non-identical label spaces such as open-set, partial and universal cross-domain continual learning problems. 

\begin{landscape}
\begin{table}[!t]
\caption{Average accuracy (\%) of all learned tasks on Office-31 dataset}\label{tab:off31}
\centering
\begin{tabular}{l|cccccc|c}
\hline
\multicolumn{1}{c|}{Method} & A→D                   & A→W                   & D→A                   & D→W                   & W→A                   & W→D                   & Avg.                 \\ \hline
AGLA                        & 55.83 ± 5.98          & 49.81 ± 3.77          & 29.61 ± 4.08          & 62.50 ± 6.02          & 31.26 ± 1.20          & 56.85 ± 5.59          & 47.64 ± 13.93        \\
CDCL                        & 9.68                  & 10.98                 & 7.02                  & 27.97                 & 6.73                  & 28.98                 & 15.23                \\
CLAMP                       & 84.29 ± 2.9           & 80.39 ± 2.0           & 61.37 ± 1.3           & 94.85 ± 0.3           & 59.08 ± 4.2           & 98.47 ± 2.2           & 79.74 ± 16.5         \\
DANN                        & 9.03 ± 1.2            & 17.77 ± 2.8           & 2.07 ± 1.1            & 4.73 ± 3.9            & 17.09 ± 2.0           & 17.88 ± 8.2           & 11.43 ± 7.1          \\
EEIL                        & 45.51 ± 2.3           & 41.02 ± 3.9           & 40.23 ± 5.0           & 79.21 ± 2.4           & 42.76 ± 1.9           & 86.72 ± 1.9           & 55.91 ± 21.2         \\
EWC                         & 15.05 ± 2.8           & 15.09 ± 1.5           & 8.67 ± 2.7            & 20.25 ± 2.3           & 14.64 ± 3.4           & 25.65 ± 3.0           & 16.56 ± 5.8          \\
HAL                         & 33.51 ± 1.4           & 32.55 ± 3.9           & 25.62 ± 1.7           & 62.81 ± 2.8           & 29.42 ± 3.3           & 53.52 ± 2.4           & 39.57 ± 2.58         \\
iCaRL                       & 47.68 ± 3.8           & 37.87 ± 1.8           & 38.77 ± 5.2           & 79.84 ± 3.6           & 42.34 ± 1.8           & 88.60 ± 1.8           & 55.85 ± 22.4         \\
IL2M                        & 48.12 ± 2.0           & 43.18 ± 3.3           & 39.95 ± 5.4           & 79.61 ± 2.2           & 42.84 ± 1.7           & 86.46 ± 1.6           & 56.69 ± 20.7         \\
LwF                         & 19.85 ± 2.1           & 18.30 ± 1.9           & 7.67 ± 2.4            & 21.18 ± 4.8           & 16.23 ± 2.3           & 27.20 ± 2.4           & 18.41 ± 6.4          \\
MAS                         & 15.95 ± 3.9           & 14.90 ± 1.5           & 8.67 ± 2.7            & 20.25 ± 2.3           & 14.52 ± 3.4           & 25.65 ± 3.0           & 16.66 ± 5.8          \\
RWalk                       & 15.95 ± 3.0           & 15.09 ± 1.5           & 8.61 ± 2.7            & 20.25 ± 2.3           & 14.64 ± 3.4           & 25.28 ± 3.0           & 16.64 ± 5.6          \\
SI                          & 15.59 ± 2.3           & 14.67 ± 1.9           & 8.38 ± 2.7            & 19.86 ± 2.9           & 14.91 ± 3.0           & 24.29 ± 2.3           & 16.28 ± 5.4          \\ \hline
\textbf{REFEREE}            & \textbf{93.55 ± 2.1} & \textbf{86.65 ± 2.9} & \textbf{81.86 ± 0.9} & \textbf{99.39 ± 0.5} & \textbf{82.89 ± 0.9} & \textbf{100} & \textbf{90.72 ± 1.2} \\ \hline
\end{tabular}
\end{table}

\begin{table}[!t]
\caption{Average accuracy (\%) of all learned tasks on DomainNet dataset}\label{tab:dom}
\centering
\begin{tabular}{c|cc|c}
\hline
Method           & clipart→sketch & sketch→clipart & Avg.           \\ \hline
CDCL             & 3.49           & 2.17           & 2.83           \\
CLAMP            & 12.87          & 13.67          & 13.27          \\ \hline
\textbf{REFEREE} & \textbf{28.70} & \textbf{36.62} & \textbf{32.66} \\ \hline
\end{tabular}
\end{table}

\begin{table}[p]
\caption{Average accuracy (\%) of all learned tasks on VisDA dataset}\label{tab:dig}
\centering
\begin{tabular}{c|c|c|c|c|c|c|c|c|c|c|c|c|c|c}
\hline
Method & AGLA       & CDCL  & CLAMP       & DANN        & EEIL        & EWC         & HAL         & iCaRL       & IL2M        & LwF         & MAS         & RWalk       & SI          & \textbf{REFEREE}     \\ \hline
VisDA  & 25.8 ± 2.7 & 10.16 & 30.71 ± 1.8 & 16.16 ± 0.1 & 15.99 ± 1.8 & 12.65 ± 3.5 & 16.87 ± 1.6 & 15.67 ± 1.7 & 19.58 ± 1.4 & 13.10 ± 1.2 & 13.45 ± 1.0 & 13.91 ± 1.7 & 13.40 ± 2.1 & \textbf{85.12 ± 0.4} \\ \hline
\end{tabular}
\end{table}

\begin{table}[p]
\centering
\caption{Average accuracy (\%) of all learned tasks on Office-Home dataset}\label{tab:hom}
\begin{tabular}{l|cccccccccccc|c}
\hline
\multicolumn{1}{c|}{Method} & Ar→Cl                                    & Ar→Pr                                    & Ar→Ew                                    & Cl→Ar                                    & Cl→Pr                                    & Cl→Rw                                    & Pr→Ar                                    & Pr→Cl                                    & Pr→Rw                                    & Rw→Ar                                    & Rw→Cl                                    & Rw→Pr                                     & Avg.                                     \\ \hline
AGLA                        & 18.58 ± 1.5                              & 29.89 ± 2.1                              & 34.20 ± 4.0                              & 17.84 ± 1.5                              & 23.39 ± 1.3                              & 28.57 ± 2.9                              & 20.02 ± 1.0                              & 19.19 ± 2.3                              & 32.89 ± 1.9                              & 28.67 ± 5.4                              & 20.61 ± 3.0                              & 41.17 ± 2.6                               & 25.25 ± 7.5                              \\
CDCL                        & 4.15                                     & 4.77                                     & 6.01                                     & 5.56                                     & 8.32                                     & 6.35                                     & 3.92                                     & 5.02                                     & 6.29                                     & 6.45                                     & 6.26                                     & 8.35                                      & 5.95                                     \\
CLAMP                       & 42.57 ± 0.9                              & 62.20 ± 1.0                              & 63.63 ± 1.7                              & 38.50 ± 2.0                              & 49.47 ± 1.5                              & 49.92 ± 0.9                              & 40.63 ± 3.1                              & 40.45 ± 1.0                              & 55.29 ± 1.9                              & 54.88 ± 1.8                              & 45.21 ± 1.3                              & 63.18 ± 1.8                               & 50.49 ± 9.3                              \\
DANN                        & 4.03 ± 0.8                               & 4.40 ± 1.1                               & 5.82 ± 0.6                               & 2.62 ± 3.0                               & 5.00 ± 1.7                               & 5.01 ± 1.8                               & 4.71 ± 1.2                               & 3.48 ± 1.7                               & 5.82 ± 1.0                               & 6.06 ± 1.4                               & 4.21 ± 0.6                               & 7.59 ± 0.7                                & 4.90 ± 1.3                               \\
EEIL                        & 26.70 ± 1.2                              & 44.10 ± 1.9                              & 52.80 ± 1.2                              & 17.21 ± 1.0                              & 28.64 ± 0.8                              & 31.37 ± 1.2                              & 28.35 ± 1.6                              & 20.41 ± 0.9                              & 45.52 ± 1.1                              & 36.26 ± 1.6                              & 25.83 ± 1.4                              & 51.74 ± 0.9                               & 34.08 ± 11.9                             \\
EWC                         & 7.34 ± 1.6                               & 11.68 ± 1.5                              & 15.38 ± 1.4                              & 5.06 ± 1.6                               & 7.70 ± 0.6                               & 7.51 ± 0.7                               & 7.68 ± 1.4                               & 6.42 ± 0.7                               & 10.35 ± 0.6                              & 13.78 ± 1.5                              & 7.75 ± 1.8                               & 13.43 ± 0.2                               & 9.51 ± 3.3                               \\
HAL                         & 10.90 ± 1.4                              & 20.34 ± 1.4                              & 29.51 ± 1.3                              & 10.17 ± 1.7                              & 13.04 ± 1.9                              & 15.72 ± 1.9                              & 14.77 ± 2.3                              & 12.38 ± 1.6                              & 26.32 ± 1.4                              & 20.50 ± 2.4                              & 13.34 ± 1.3                              & 31.13 ± 1.0                               & 18.18 ± 1.6                              \\
iCaRL                       & 28.03 ± 1.0                              & 44.93 ± 1.3                              & 53.91 ± 2.1                              & 19.81 ± 1.0                              & 30.43 ± 2.0                              & 33.11 ± 1.3                              & 30.28 ± 2.1                              & 23.39 ± 1.5                              & 46.09 ± 2.1                              & 38.72 ± 1.4                              & 25.93 ± 1.4                              & 52.40 ± 0.9                               & 35.59 ± 11.4                             \\
IL2M                        & 26.60 ± 1.0                              & 43.82 ± 1.8                              & 52.72 ± 1.1                              & 18.03 ± 0.8                              & 28.14 ± 1.2                              & 30.85 ± 1.5                              & 28.17 ± 2.5                              & 20.24 ± 0.8                              & 46.12 ± 1.5                              & 37.52 ± 1.5                              & 25.85 ± 1.8                              & 52.08 ± 1.0                               & 34.18 ± 12.0                             \\
LwF                         & 8.11 ± 2.1                               & 11.91 ± 1.4                              & 16.20 ± 1.3                              & 5.09 ± 2.6                               & 8.68 ± 0.8                               & 9.40 ± 1.1                               & 8.30 ± 1.6                               & 7.35 ± 1.2                               & 12.65 ± 1.7                              & 13.56 ± 1.3                              & 8.51 ± 1.7                               & 15.02 ± 0.7                               & 10.40 ± 3.4                              \\
MAS                         & 7.38 ± 2.1                               & 11.59 ± 1.0                              & 15.51 ± 2.0                              & 4.69 ± 1.8                               & 7.42 ± 1.1                               & 7.12 ± 0.7                               & 7.81 ± 1.9                               & 6.44 ± 1.0                               & 10.87 ± 0.7                              & 14.09 ± 2.1                              & 7.46 ± 1.7                               & 13.37 ± 0.4                               & 9.48 ± 3.5                               \\
RWalk                       & 7.18 ± 1.8                               & 11.53 ± 1.4                              & 15.25 ± 1.5                              & 5.15 ± 2.2                               & 7.67 ± 0.6                               & 7.45 ± 0.4                               & 7.53 ± 1.7                               & 6.54 ± 1.0                               & 10.91 ± 0.6                              & 13.56 ± 1.8                              & 7.75 ± 1.6                               & 13.53 ± 0.4                               & 9.50 ± 3.3                               \\
SI                          & 7.44 ± 1.7                               & 11.15 ± 1.0                              & 15.50 ± 1.4                              & 4.28 ± 1.9                               & 7.59 ± 0.8                               & 7.05 ± 1.0                               & 7.32 ± 1.2                               & 6.35 ± 1.3                               & 9.92 ± 1.0                               & 12.85 ± 1.4                              & 7.87 ± 1.8                               & 13.73 ± 0.4                               & 9.25 ± 3.4                               \\ \hline
\textbf{REFEREE}            & \multicolumn{1}{l}{\textbf{39.17 ± 2.2}} & \multicolumn{1}{l}{\textbf{74.57 ± 0.9}} & \multicolumn{1}{l}{\textbf{82.94 ± 1.2}} & \multicolumn{1}{l}{\textbf{65.88 ± 2.9}} & \multicolumn{1}{l}{\textbf{70.74 ± 0.4}} & \multicolumn{1}{l}{\textbf{75.03 ± 1.4}} & \multicolumn{1}{l}{\textbf{60.32 ± 2.5}} & \multicolumn{1}{l}{\textbf{30.26 ± 1.2}} & \multicolumn{1}{l}{\textbf{80.94 ± 1.3}} & \multicolumn{1}{l}{\textbf{64.76 ± 1.9}} & \multicolumn{1}{l}{\textbf{33.84 ± 1.7}} & \multicolumn{1}{l|}{\textbf{77.22 ± 1.3}} & \multicolumn{1}{l}{\textbf{62.97 ± 1.6}} \\ \hline
\end{tabular}
\end{table}
\end{landscape}

% or


%\appendices

%\section{}

%\section*{Acknowledgment}


\ifCLASSOPTIONcaptionsoff
  \newpage
\fi



%\bibliographystyle{IEEEtran}
%\begin{thebibliography}{1}


%\end{thebibliography}

\bibliographystyle{IEEEtran}
\bibliography{references}


%\end{IEEEbiography}


%\newpage



%\vfill


\end{document}